\documentclass{article}

\usepackage{arxiv}

\usepackage[utf8]{inputenc} 
\usepackage[T1]{fontenc}    
\usepackage{hyperref}       
\usepackage{url}            
\usepackage{booktabs}       
\usepackage{amsfonts}       
\usepackage{nicefrac}       
\usepackage{microtype}      
\usepackage{lipsum}		
\usepackage{graphicx}
\usepackage{natbib}
\usepackage{doi}

\title{Exploring Social Desirability Response Bias in Large Language Models: 
Evidence from GPT-4 Simulations}

\author{
    \begin{minipage}[t]{0.32\textwidth}
        \centering
        \href{https://orcid.org/0000-0001-9204-0867}{\includegraphics[scale=0.06]{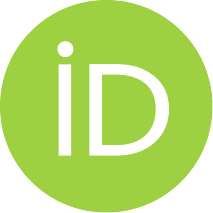}\hspace{1mm}Sanguk ~Lee}\thanks{These authors contributed equally to this work.} \\
        Department of Communication Studies\\
        Texas Christian University\\
        Fort Worth, TX 76219 \\
        \texttt{sanguk.lee@tcu.edu}
    \end{minipage}
    \hfill
    \begin{minipage}[t]{0.32\textwidth}
        \centering
        Kai-Qi ~Yang\footnotemark[1] \\
        Department of Computer Science and Engineering\\
        Michigan State University\\
        East Lansing, MI 48824 \\
        \texttt{kqyang@msu.edu}
    \end{minipage}
    \hfill
    \begin{minipage}[t]{0.32\textwidth}
        \centering
        \href{https://orcid.org/0000-0002-2588-7491}{\includegraphics[scale=0.06]{orcid.pdf}\hspace{1mm}Tai-Quan ~Peng}\thanks{Corresponding Author: \texttt{pengtaiq@msu.edu}} \\
        Department of Communication\\
        Michigan State University\\
        East Lansing, MI 48824 \\
        \texttt{pengtaiq@msu.edu}
    \end{minipage}
    \vspace{1cm} 
    \\
    \begin{minipage}[t]{0.45\textwidth}
        \centering
        \href{https://orcid.org/0000-0001-5099-2309}{\includegraphics[scale=0.06]{orcid.pdf}\hspace{1mm}Ruth ~Heo} \\
        Department of Communication\\
        Michigan State University\\
        East Lansing, MI 48824 \\
        \texttt{heoruth@msu.edu}
    \end{minipage}
    \hfill
    \begin{minipage}[t]{0.45\textwidth}
        \centering
        Hui ~Liu \\
        Department of Computer Science and Engineering\\
        Michigan State University\\
        East Lansing, MI 48824 \\
        \texttt{liuhui7@msu.edu}
    \end{minipage}
}

\date{\today}

\renewcommand{\shorttitle}{Lee et al. (2024) Exploring SDR Bias in LLMs}

\hypersetup{
    pdftitle={Exploring Social Desirability Response Bias in Large Language Models},
    pdfsubject={q-bio.NC, q-bio.QM},
    pdfauthor={Sanguk Lee, Kai-Qi Yang, Tai-Quan Peng, Ruth Heo, Hui Liu},
    pdfkeywords={Social desirability response bias, large language models, cimmitment statement, survey simluation}
}

\begin{document}
\maketitle

\begin{abstract}
	Large language models (LLMs) are employed to simulate human-like responses in social surveys, yet it remains unclear if they develop biases like social desirability response (SDR) bias. To investigate this, GPT-4 was assigned personas from four societies, using data from the 2022 Gallup World Poll. These synthetic samples were then prompted with or without a commitment statement intended to induce SDR. The results were mixed. While the commitment statement increased SDR index scores, suggesting SDR bias, it reduced civic engagement scores, indicating an opposite trend. Additional findings revealed demographic associations with SDR scores and showed that the commitment statement had limited impact on GPT-4's predictive performance. The study underscores potential avenues for using LLMs to investigate biases in both humans and LLMs themselves.
\end{abstract}


\section{Introduction}
Large language models (LLMs) are not only powerful tools for language tasks (e.g., topic extraction, sentiment analysis), but also hold significant potential for creative applications beyond traditional linguistic boundaries. Their remarkable ability to analyze and generate human-like text across diverse contexts has opened new avenues for innovation. One particularly important application of LLMs lies in their potential to simulate human responses in social surveys \cite{Argyle2023, Lee2024, Qu2024} and experiments \cite{Hewitt2024, Mei2024}, offering a promising alternative for observing and understanding attitudes, perceptions, and behaviors. While debates persist regarding the nature of language—whether it primarily serves as a tool for communication or thought \cite{Fedorenko2024}--these discussions should not deter social scientists from conducting empirical investigations into the capacity of LLMs to simulate human responses in survey contexts. \par

Using LLMs to simulate human responses provides substantial advantages for social science research. LLMs can offer a quick, cost-effective way to generate data that can refine survey designs and enhance sample representativeness, especially for underrepresented social groups. They can also fill in missing data and explore new possibilities by simulating responses that may not have been captured in the original survey \cite{Kim2023}. These features make LLMs a powerful tool for achieving more comprehensive and accurate analyses, thereby having the potential to advance social science research. \par

However, when we embrace LLMs in social science research, it is important to comprehensively assess LLMs’ ability to imitate a wide range of human aspects including perceptions, behaviors, and biases. While many studies have focused on investigating LLMs’ ability to simulate human beliefs \cite{Lee2024}, perceptions \cite{Hwang2023}, and behaviors \cite{Aher2023, Argyle2023}, one critical area that has received less attention is whether LLMs can inherit human biases commonly discussed in survey research. Such human bias possibly streams down to LLMs due to the inherent bias embedded in the training and human-feedback data \cite{Bai2022, Pfohl2024}. 

\section{Understanding Social Desirability Response Bias in Large Language Models}
\label{sec:headings}
Investigating bias in LLMs in their applications in social surveys is crucial. First, evaluating how well LLMs replicate biases inherent in survey designs is critical for accurately simulating human behavior and improving the reliability of these simulations. Second, unnoticed biases in LLMs, whether stemming from the models themselves or arising during their applications, can lead to flawed analysis, perpetuate harmful stereotypes, and compromise both future survey design and the integrity of AI technologies. By understanding whether specific biases exist in LLMs and how they align with or differ from human biases, researchers can better interpret survey outcomes and enhance the utility of LLMs in social research. \par

Within the spectrum of biases pervasive in social surveys, social desirability response (SDR) bias \cite{Crowne1960} remains a long-standing concern for social scientists. SDR bias is characterized by respondents’ inclination to provide answers that make them look good. Rooted in the tendency to conform to social expectations \cite{Schlenker1989}, SDR bias has been widely observed and discussed in survey research across various data collection modes \cite{Dodou2014, Holbrook2003, Kreuter2008}. Survey questions on sensitive topics like politics, ethics, or behaviors are particularly susceptible to the SDR bias \cite{Johnson2003, Krumpal2013, Streb2008}. \par

To collect truthful responses in social surveys, social scientists have tried multiple strategies. One effective strategy is to include a warning message in social surveys, which aims to encourage respondents to follow instructions or to think carefully \cite{Huang2012, Krosnick2000}. However, while warning messages in a survey have been found to increase respondents’ motivation to generate truthful answers to survey questions, they can also induce reputation-management concerns that manifest themselves in more socially desirable attitudes and behaviors \cite{Clifford2015}. In other words, the presence of a warning message in the survey questionnaire plays a double-edged sword that can elevate the respondents’ motivation or their SDR bias. \par

This current research investigates if the SDR bias will manifest in responses from LLMs when they are applied in the survey context. We use the commitment statement as a potential stimulus that induces social desirability bias in the responses of LLMs. The commitment statement, as a type of warning message, is known to provoke SDR bias by emphasizing participants’ commitments in responding to survey questions \cite{Clifford2015}. This heightened engagement can make respondents more conscious of how they present themselves, leading them to alter their answers to align with socially acceptable norms. \par

Comparing LLMs’ responses between the two conditions, with and without the commitment statements, this study examines the presence and patterns of SDR in LLMs. We investigate this phenomenon using two different SDR indicators: SDR index and civic engagement (CE, hereafter) index. Both SDR index and civic engagement activity have been used to capture SDR bias in previous studies \cite{Clifford2015, Holbrook2010}. SDR index is specifically designed to measure the extent of social desirability bias in respondents, directly capturing the tendency to provide socially desirable answers rather than truthful ones. CE index, on the other hand, intends to measure actual behaviors related to civic participation, such as election participation, charity donations, and helping strangers. Although CE index is not specifically designed to measure SDR bias, the responses it captures can still be influenced by it \cite{Holbrook2010}. Assuming that a commitment statement leads to SDR bias among human participants, it is hypothesized that:  

\textit{H1a--b: Synthetic samples produced by LLMs exposed to a commitment statement will display social desirability bias, resulting in higher levels of a) SDR scores and b) CE scores compared to synthetic samples not exposed to a commitment statement.}

This study further examines the association between demographics and SDR bias. Previous studies have found that certain demographic groups exhibit a higher tendency toward SDR bias than others. For example, SDR bias has been positively associated with age, with older individuals displaying higher levels of SDR bias compared to younger individuals \cite{Ausmees2022, Dijkstra2001, Soubelet2011}. Similarly, better-educated groups have been shown to exhibit greater SDR bias than less-educated groups \cite{Clifford2015}. In addition to the main effect of demographics on SDR bias, a moderation effect between demographics and survey design was observed. Specifically, the impact of a commitment statement on SDR bias was found to be greater among better-educated individuals \cite{Clifford2015}. This study investigates whether these patterns persist in synthetic samples generated by LLMs.

\textit{RQ1: How are demographics of synthetic samples associated with SDR bias?}

\textit{RQ2: Does the effect of a commitment statement on SDR bias vary across demographic groups?}

Additionally, the study examines if SDR bias induced by a commitment statement will influence the predictive performance of LLMs. LLMs with the commitment statement compared to those without the commitment statement may generate responses that misalign with real human responses due to the induced SDR bias. This will improve our understanding of downstream effect of SDR bias and the role of a commitment statement in the predictive performance of LLMs. 

\textit{RQ3: Does a commitment statement affect the predictive performance of GPTs?}

\section{Methods}
\label{sec:others}
\subsection{Benchmark Data}
This study used the dataset collected in the 2022 Gallup World Poll as the benchmark. The Gallup World Poll is a large-scale survey study conducted annually in over 140 societies, representing more than 99\% of the world's adult population. It collects data on a wide range of topics such as Internet access, civic engagement, leadership approval, media freedom, and many other subjects related to human needs and potential. This study focuses on four societies: Hong Kong, South Africa, the United Kingdom, and the United States. Our focus on these four societies is based on their commonalities and differences. English serves as a primary language in all of them, ensuring a consistent linguistic context. However, their distinct cultural backgrounds offer an opportunity to explore how these cultural differences may influence LLMs’ responses with regard to SDR bias. \par

For each society under study, we randomly selected 500 respondents from the Gallup World Poll sample using stratified sampling based on age and gender. Due to some synthetic respondents generated by GPT providing invalid answers, the final valid sample sizes are as follows: 447 for Hong Kong, 440 for South Africa, 437 for the United Kingdom, and 454 for the United States.

\subsection{Prompt Setting and Research Design}
We employed GPT-4 (version \textit{gpt-4-1106-preview}), recognized as one of the most powerful LLMs, to generate synthetic individuals and their responses. To encourage the variance of responses, the temperature is set as 1.5. To create realistic personas, we constructed detailed prompts that included three sets of variables from the Gallup World Poll. The first set of variables are demographic information provided by respondents in the survey, including age, gender, marital status, education level, annual household income, employment status, household size, perceived economic circumstances. The second set of variables are survey contexts, including geolocation, year, and date of the interview. The third set of variables is the Community Basics Index consisting of seven items, including respondents’ perceptions about the public transportation system, roads and highways, air quality, and more. The Community Basics Index offers insights into respondents' social and physical surroundings. The Gallup World Poll survey items used in this study are available at Gallup Worldwide Research Methodology and Codebook \url{https://news.gallup.com/poll/165404/world-poll-methodology.aspx}. An example prompt is as follows: \par

“I am [age] years old. My current marital status is [marital status]. As for education, I have [education level]. My current employment status is [employment status]. My annual household income is [annual household income]. I am from [urbanity]. My HHSIZE (Total Number of People Living in Household) is [HHSIZE]. \par
I am residing in [country]. Today is [year and date].
In the city or area where I live, I am [public transportation] with the public transportation systems. \par
In the city or area where I live, I am [road quality] with the roads and highways. In the city or area where I live, I am [education quality] with the educational system or the schools. In the city or area where I live, I am [air quality] with the quality of air. In the city or area where I live, I am [water quality] with the quality of water. In the city or area where I live, I am [quality healthcare] with the availability of quality healthcare. In the city or area where I live, I am [affordable housing] with the availability of good affordable housing.” \par

This study employed a within-subject experiment design where synthetic individuals were exposed to both the presence and absence of the commitment statement. The commitment message reads “It is important to us that participants in our survey pay close attention to the materials. Are you willing to carefully read the materials and answer all of the questions to the best of your ability?” After the treatment, synthetic individuals were prompted to provide their answers to targeted questions. Due to limitations in instruction-following abilities \cite{Qin2024} and the occurrence of hallucinations \cite{Huang2023}, LLMs do not always adhere to instructions, often producing responses that do not align with the predefined answer options. To mitigate this issue, we prompted them to answer each set of questions independently. For instance, after the pre-setup, synthetic individuals responded to a set of SDR items. We then started over the pre-setup to have them respond to civic engagement activity items. Examples of prompts are available in the supplemental document. 

\subsection{Measurements from Synthetic Samples}
\subsubsection{Social Desirability Response (SDR) Index}
The SDR index was derived from \citep{Ballard1992}. This index consists of 13 true/false items, with seven statements describing socially (un)desirable but uncommon behaviors and the other six representing socially (un)desirable but common behaviors. For example, the statement “I sometimes feel resentful when I don’t get my own way” illustrates an uncommon behavior. Answering “false” to this question is considered a socially desirable response. Another statement, “I am always courteous, even to people who are disagreeable,” represents a common behavior. Answering “true” to this question is considered socially desirable. The items are dichotomously scored, with a “1” indicating a socially desirable response—“true” for socially desirable items and “false” for socially undesirable ones. The SDR index score generated by synthetic samples was computed by summing the number of socially desirable responses, resulting in a score ranging from 0 to 13 (\textit{M} = 5.05, \textit{SD} = 2.88).
\subsubsection{Civic Engagement (CE) Index}
The CE index was derived from the measurements in the Gallup World Poll survey. This index consists of three items assessing civic activities in the past month, including donating money to charity, volunteering time to an organization, and helping a stranger. For instance, to measure donations to charity, an item asking “Have you done any of the following in the past month? How about donated money to a charity?” was used. A binary scale “Yes” or “No” was used for each item. The CE index score generated by synthetic samples was computed by summing the number of civic engagement activities, resulting in a score ranging from 0 to 3 (\textit{M} = 1.08, \textit{SD} = .60). 
\subsubsection{Analytic Plans}
To investigate Ha-b, we used multilevel regressions to account for the within-subject effects of synthetic individuals, with the treatment condition serving as the independent variable. To examine RQ1 and RQ2, we conducted a multilevel regression in which demographics, the treatment condition, and their interaction terms served as independent variables, while accounting for the within-subject effects of synthetic individuals. For RQ3, we computed the F1 score for each civic activity. The F1 score can be calculated for each label (e.g., "Yes" or "No"). For simplicity, we reported the F1 score of the label that produced the higher value. For charity donations and volunteering, higher F1 scores were associated with the "No" response, while for helping a stranger, higher F1 scores were associated with the "Yes" response. Thus, we reported these F1 scores. 

\section{Results}
\subsection{Reporting Socially Desirable? Effect of Commitment Statement on SDR bias (H1a--b)}
\begin{figure}
    \centering
    \includegraphics[width=0.5\linewidth]{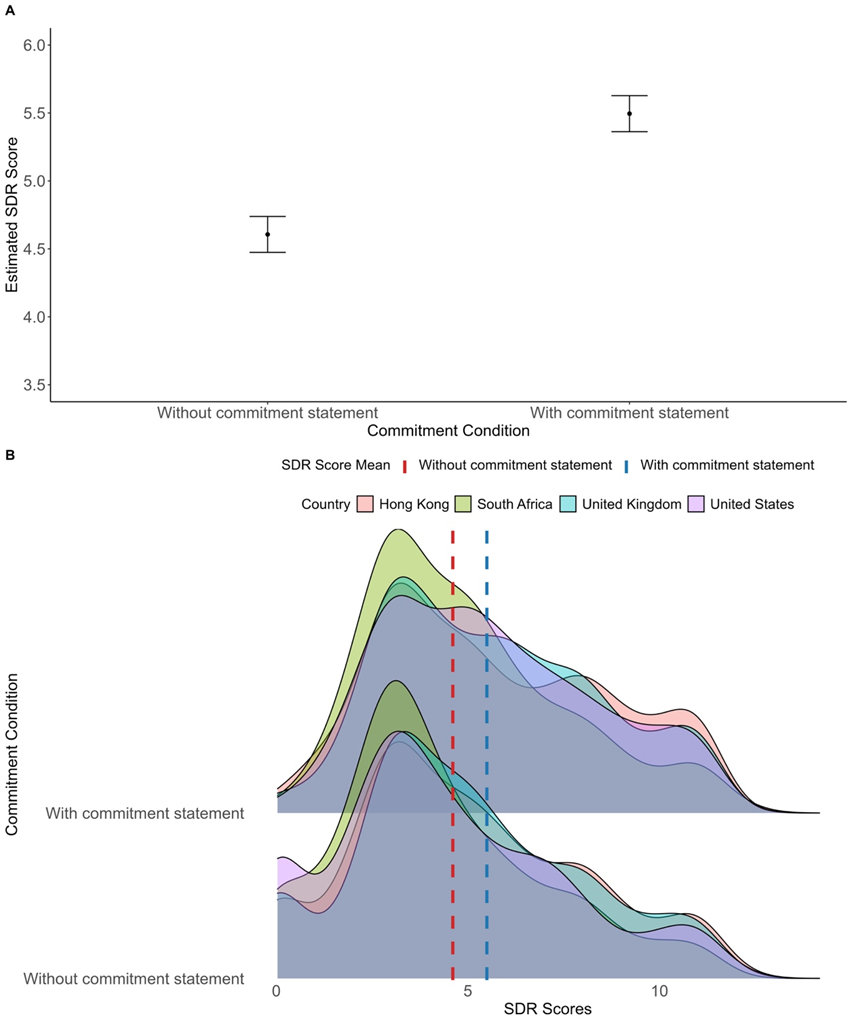}
    \caption{A) Estimated SDR Scores by Commitment Statement Condition across All Societies and B) Distribution of SDR Scores by Commitment Statement Condition for Each Society}
    \label{fig:fig1}
\end{figure}

To assess whether the commitment statement induces social desirability response bias in LLMs, we examined two scales generated by GPT-4: the SDR index and the CE index. First, we examined this phenomenon based on the SDR index. As Figure~\ref{fig:fig1}A illustrates, it is evident that the commitment statement facilitates SDR bias. SDR index was significantly higher when the commitment statement was present (estimated mean (hereafter, \textit{M}) = 5.49, \textit{SE} = 0.07, \textit{CI} 95\% = [5.36, 5.63]) than when such statement was absent (\textit{M} = 4.60, \textit{SE} = 0.07, \textit{CI} 95\% = [4.47, 4.74]). Figure~\ref{fig:fig1}B illustrates the distribution of SDR index with and without the commitment statement across societies. The distribution pattern was homogenous across the four societies, indicating that the synthetic sample exposed to the commitment statement produced higher scores than those without it regardless of the geographical context of the study. 

\begin{figure}
     \centering
     \includegraphics[width=0.5\linewidth]{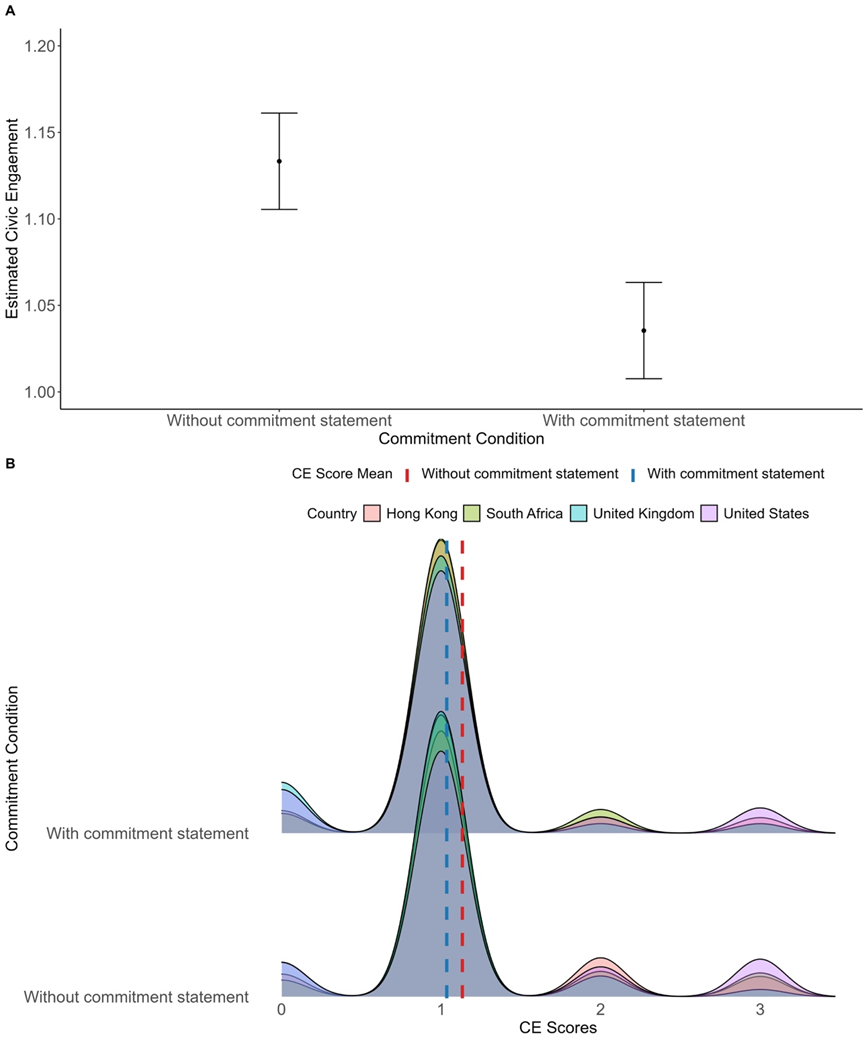}
     \caption{A) Estimated CE by Commitment Statement Condition across All Societies and B) Distribution of CE Scores by Commitment Statement Condition for Each Society}
     \label{fig:fig2}
 \end{figure}

 We investigated whether the same pattern, where SDR bias is greater in the presence of a commitment statement than in its absence, would also be observed with the CE index. Unexpectedly, we found a pattern opposite to our prediction. As Figure~\ref{fig:fig2}A illustrates, the synthetic sample with the commitment statement had lower scores of CE index (\textit{M} = 1.04, \textit{SE} = 0.01, \textit{CI} 95\% = [1.01, 1.06]) than the sample without the commitment statement (\textit{M} = 1.13, \textit{SE} = 0.01, \textit{CI} 95\% = [1.11, 1.16]). In other words, the synthetic sample with the commitment statement exhibited less SDR bias than the sample without it. The correlation between SDR index and CE index was not statistically significant (\textit{r} = .02, \textit{p} = .22), indicating these two variables were independent.

\subsection{Main and Moderation Effects of Demographics on SDR Index (RQ1 \& RQ2)}
We examined the associations between demographic characteristics and SDR bias. Our analysis focused on the SDR index, as the SDR bias was exclusively observed in this measure. This investigation revealed significant main effects of age and education on the SDR index. Specifically, synthetic individuals with older ages tended to display higher SDR scores, \textit{b} = .03, \textit{SE} = .004, \textit{t}(3461.56) = 7.47, \textit{p} < .001. Similarly, synthetic individuals with higher levels of education exhibited greater SDR scores, \textit{b} = .29, \textit{SE} = .11, \textit{t}(3517.71) = 2.54, \textit{p} < .05. Other demographics such as gender and income were not significantly associated with the SDR index. \par

We further examined whether demographic characteristics interact with the commitment statement in influencing SDR index. The impact of the commitment statement on the SDR index was positively moderated by age, \textit{b} = .01, \textit{SE} = .005, \textit{t}(1773) = 2.61, \textit{p} < .01. In other words, the presence of a commitment statement significantly increased SDR index as age increased, whereas such increase was relatively modest in the absence of a commitment statement. Other demographic variables, including education, gender, and income, did not have such significant interaction effects. 
\subsection{Limited Effect of Commitment Statement on the Predictive Performance of GPT-4 (RQ3)}
\begin{figure}
    \centering
    \includegraphics[width=0.5\linewidth]{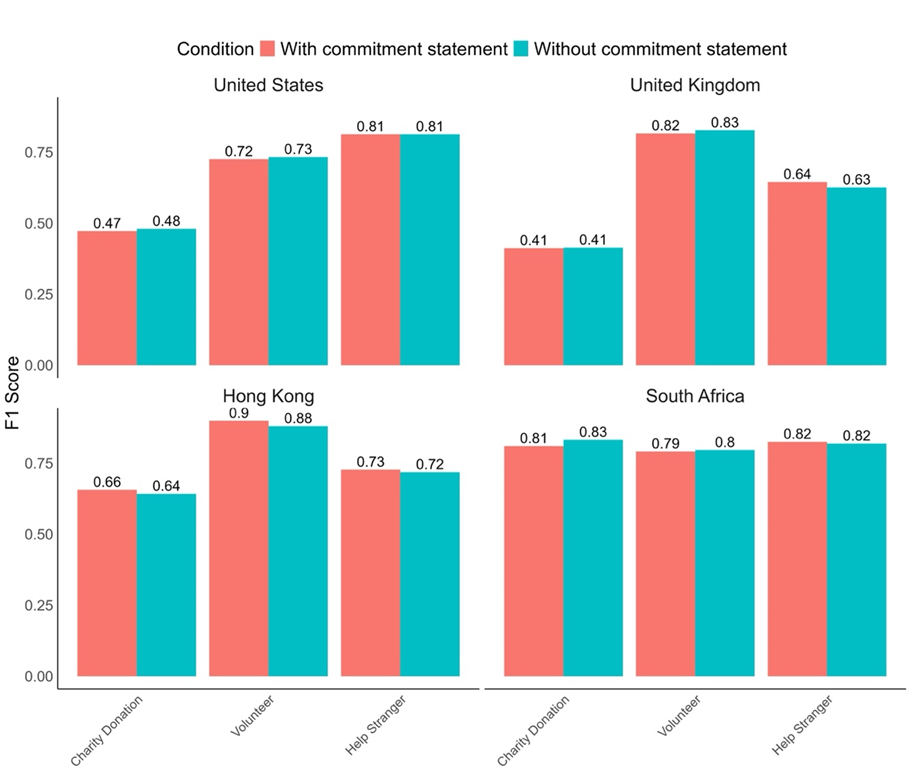}
    \caption{Comparison of F1 Scores for Civic Engagement Activities between Commitment Statement Conditions. Note. Macro-F1 scores were reported for Overall Civic Engagement. For Charity Donation and Volunteer, F1 scores were computed based on “No” response. For Help Stranger, F1 scores were computed based on “Yes” response.}
    \label{fig:fig3}
\end{figure}

Next, we examined whether the commitment statement affects the predictive performance of GPT-4. The F1 score, which measures the balance between precision and recall, was used to evaluate this performance. Figure~\ref{fig:fig3} summarizes the F1 scores for civic engagement activities between the two conditions across the four societies. Overall, the effect of the commitment statement on the predictive performance of GPT-4 was limited. Across the three civic engagement activities, F1 scores were similar between the two conditions, and this pattern was consistent across the four societies.

\begin{figure}
    \centering
    \includegraphics[width=0.5\linewidth]{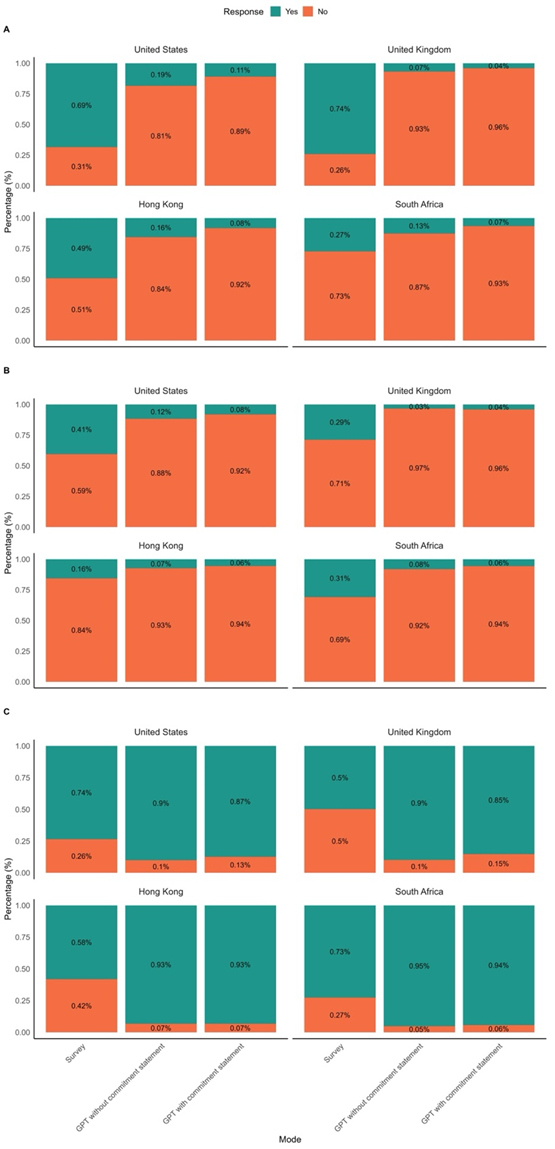}
    \caption{Stacked Bar Plots of Survey and Synthetic Sample Responses Regarding A) Charity Donations, B) Volunteer Time to Organizations, and C) Helping Strangers Across Four Societies.}
    \label{fig:fig4}
\end{figure}

Figure~\ref{fig:fig4} illustrates the response patterns of the survey and synthetic samples on civic activities across the four societies. Comparing these response patterns provides more context about the performance of GPT-4. Two key patterns emerge. First, the response patterns from synthetic individuals were nearly identical across societies. Synthetic samples predominantly responded ‘No’ to both the charity donation and volunteer questions, while answering ‘Yes’ to the help stranger question. Second, GPT-4’s predictive performance was particularly lower in the USA and UK, as its predictions deviated more from the actual responses of individuals in these societies than Hong Kong and South Africa.

\section{Discussion and Conclusions}
This study investigates social desirability response (SDR) bias in large language models (LLMs). Using GPT-4 and the 2022 Gallup World Poll data, we created synthetic samples from four different societies including Hong Kong, South Africa, USA, and UK. 
\subsection{Do LLMs Acquire Social Desirability Bias? An Unresolved Question that Needs Further Attention}
Our findings did not provide consistent evidence of SDR bias in LLMs when simulating human responses to survey questions. On one hand, the presence of a commitment statement increased scores on the SDR index, indicating potential SDR bias. On the other hand, it lowered scores on the civic engagement (CE) index, suggesting an opposite pattern from what is typically associated with SDR bias. \par

This inconsistency in our study suggests that a deeper understanding of SDR bias in LLMs is needed. A dual role of warning messages has been documented in the literature. While commitment statements can unintentionally induce SDR bias \cite{Clifford2015}, they can also enhance truthful responses by increasing participants' attention to the survey \cite{Hibben2022}. Our findings suggest that the effect of a commitment statement may vary depending on the nature of the survey items \cite{Krumpal2013}. The SDR index presents hypothetical situations, allowing respondents to adjust their answers in socially desirable ways. In contrast, while individuals might exaggerate their engagement in civic behaviors to appear more socially acceptable, doing so would involve lying, which could provoke discomfort and guilt. This tendency to report honestly on the CE index might be stronger in the presence of a commitment statement. GPT-4 might capture this nuanced aspect of human behavior and reflected the contradictory pattern through synthetic samples. \par

SDR bias is a complex phenomenon. It can be unconsciously encouraged through self-deception or strategically motivated by impression management \cite{Paulhus1984}. It can also manifest in various forms, such as overreporting positive behaviors, underreporting negative behaviors, altering attitudes on controversial issues, and supporting mainstream opinions \cite{Holtgraves2017}. Although our study does not provide clear evidence of SDR bias in LLMs, this single study is insufficient to conclude that LLMs will not exhibit such bias when simulating human responses in social surveys. Given the wide range of influences on SDR, from question framing to survey mode \cite{Heerwegh2009, Kreuter2008, Tourangeau2007}, future research should explore how other stimuli may trigger SDR in LLMs, broadening our understanding of how these models respond to diverse survey contexts.
\subsection{Associations between Demographics and SDR Bias}
GPT-4 captured the associations between demographics and SDR bias. Consistent with findings from human research, older and better-educated synthetic individuals exhibited greater SDR bias compared to their counterparts \cite{Ausmees2022, Clifford2015, Dijkstra2001, Soubelet2011}. These findings suggest that GPT-4 mirrors how such bias interacts with demographic factors. Several studies have demonstrated that LLMs can integrate demographic information to estimate perceptions and behaviors \cite{Argyle2023, Lee2024}. This study contributes to this line of research by extending its scope to more complex and subtle psychological phenomena such as SDR bias. \par

In addition to these main effects, our analysis revealed a moderation effect of demographics on the commitment statement. Specifically, the commitment statement had a more pronounced effect on increasing SDR scores among older synthetic personas. This novel finding, not previously explored in human research, highlights the potential of LLMs for social science exploration and suggests that similar results could be replicated with human participants. A possible explanation is that older adults may be more motivated by social expectations and norms, which can lead them to present themselves in a more socially desirable manner. The commitment statement likely heightened their awareness of social expectations, thereby amplifying their tendency to align their responses with perceived social acceptability. 
\subsection{Impact of SDR Bias on Predictive Performance}
Overall, GPT-4 demonstrates a satisfactory level of performance in estimating human civic activities. Although GPT-4’s responses to civic activities systematically vary based on the condition of the commitment statement, this effect is marginal in altering the predictive performance of GPT-4. However, a closer investigation reveals GPT-4’s lack of sensitivity to cultural variations. In particular, there was a greater discrepancy between human and synthetic samples from the USA and the UK compared to those from Hong Kong and South Africa in responses to charity donations. While many human participants from the USA and UK reported that they donated their money to charity, GPT-4 underestimated this generosity. \par

These results are intriguing, especially considering recent findings on Western-centric biases in LLMs \cite{Naous2024, Qu2024}. One hypothesis is that GPT-4 may be overcompensating for perceived biases by underrepresenting prosocial behaviors in Western societies like the USA and UK. This could stem from efforts to correct for overrepresentation of Western norms in training data, inadvertently leading to an underestimation of behaviors such as charitable giving, which are culturally significant in these regions. Another theoretical framework that might explain this discrepancy is the Cultural Tightness-Looseness Theory \cite{Gelfand2011}, which posits that societies vary in the strength of their social norms and tolerance for deviant behavior. GPT-4 may not accurately capture these cultural nuances, leading to homogenized responses. Additionally, the model’s training data may lack sufficient cultural specificity, causing it to generalize behaviors across societies without accounting for regional differences. Further investigation into the training data’s composition and the model’s cultural sensitivity is necessary to understand and address these predictive inaccuracies.
\subsection{Methodological Implications}
Overall, our data provides inconclusive results about SDR bias in LLMs. The lack of empirical evidence from human data further obscures our understanding of this bias in LLMs. With only a handful of studies examining the effects of warning messages on SDR bias, it remains unclear whether our synthetic data reflects SDR bias in the same way as human data. Although SDR bias has been studied for decades, its numerous influential factors—such as survey design \cite{Clifford2015}, items \cite{Holbrook2010}, modes \cite{Holbrook2003}, and individual characteristics \cite{Crowne1960}--complicate its investigation. \par
Using GPT-4 simulations presents significant potential for social science research by allowing us to investigate underexplored aspects of human bias more quickly and cost-effectively. LLMs’ ability to replicate human behaviors and biases has been increasingly documented \cite{Argyle2023, Lee2024, Qu2024}, suggesting that our synthetic data may reflect the pattern of human SDR bias to some extent. Our findings offer plausible scenarios related to SDR bias that may merit further investigation. This LLM research, inspired by human subject studies, has the potential to inspire future research involving human subjects to investigate previously unknown patterns of SDR bias. Such a symbiotic research approach can deepen our understanding of both human and LLM behaviors and biases.

\bibliographystyle{unsrtnat}
\bibliography{references}  






\end{document}